\documentclass[runningheads]{llncs}
\usepackage[T1]{fontenc}
\usepackage{graphicx}
\newcommand{\best}[1]{\textbf{#1}}
\usepackage{booktabs}
\usepackage{multirow}
\usepackage{array}
\usepackage{amsmath}
\usepackage{xcolor}
\usepackage{graphicx}
\usepackage{caption}
\usepackage{float}
\usepackage{tabularx}
\usepackage{adjustbox}
\usepackage{makecell}
\usepackage[numbers]{natbib}

\begin{document}
\title{EmoMed: An Emotionally-Aware Agent for Multimodal Medical Support with Real-Time Information Retrieval}
\author{
    Ivan Nasonov\inst{1}\orcidID{0009-0003-3292-8479} \and
    Nikita Glazkov\inst{2,3}\orcidID{0000-0003-0400-9849}\thanks{Nikita Glazkov - Corresponding author (glazkov@airi.net).} \and
    Ivan Makovetskiy\inst{3}\orcidID{0009-0000-2921-8861}
    \and 
    Mikhail Mozikov\inst{1,2}\orcidID{0000-0003-0594-867X} \and
    Daniil Sukhorukov\inst{2}\orcidID{0009-0005-1456-4784} \and
    Andrey Savchenko\inst{1,4}\orcidID{0000-0001-6196-0564} \and
    Ilya Makarov\inst{1,2,5}\orcidID{0000-0002-3308-8825} 
}
\authorrunning{I. Nasonov et al.: EmoMed: An Emotionally-Aware Agent}
\institute{ISP RAS Research Center for Trusted Artificial Intelligence, 109004 Moscow, Russia \and
AXXX, 119049 Moscow, Russia \and
National University of Science and Technology (NUST) MISIS, 119049 Moscow, Russia \and
SB AI Lab, 117997 Moscow, Russia \and
Research Center of the Artificial Intelligence Institute, Innopolis University, 420500 Innopolis, Russia \and
SkolTech, 121205 Moscow, Russia
}

\maketitle              % typeset the header of the contribution
\begin{abstract}
We present EmoMed -- a multimodal medical consultation agent that adapts its responses based on users' emotional states while maintaining clinical accuracy. The system processes text and medical images, detects affect indicators (anxiety, confusion, urgency) from user input, and adjusts response tone, structure, and detail level accordingly. To ensure factual reliability, the agent grounds clinical information through a dual retrieval mechanism: web-based fact-checking and an API-connected, continuously updated medical knowledge base. We evaluate our approach across seven state-of-the-art language models (GPT-4/5, Qwen3, Llama 4, Gemini 2.5, Grok4, Claude3) using comprehensive metrics including LLM-as-judge assessments, MedQA style accuracy tests, BERT Score, safety/helpfulness ratings, and multimodal medical benchmarks. The results demonstrate that emotionally adaptive responses consistently outperform neutral baseline across evaluation dimensions, without compromising clinical accuracy. A controlled user study validated these findings, with participants reporting improved perceived empathy and communication clarity, while maintaining trust in factual accuracy.

Source code: \url{https://github.com/NasonovIvan/EmoMed-Agent}


\end{abstract}

\section{Introduction}

Large language models (LLMs) are rapidly entering clinical and patient-facing workflows for education, pre-consult triage, reference Q\&A, decision support, summarization, and note generation \citep{car2020conversational,barreda2025transforming,liu2025reviewApplyingLLMHealthcare}. Several studies report strong performance on medical knowledge and reasoning benchmarks (USMLE, MedQA / MultiMedQA) \citep{singhal2022nature,nori2023gpt4capabilities,singhal2023medpalm}, yet the community stresses the need for more rigorous, realistic, and transparent evaluations to ensure safe deployment in high-stakes settings \citep{alwakeel2025rigor}. Persistent risks include hallucinations, uneven domain coverage, and knowledge drift in dynamic medical contexts \citep{kim2025medicalHallucinations,li2025lokisDance}.

\begin{figure*}[t]
\centering
\includegraphics[width=0.8\textwidth]{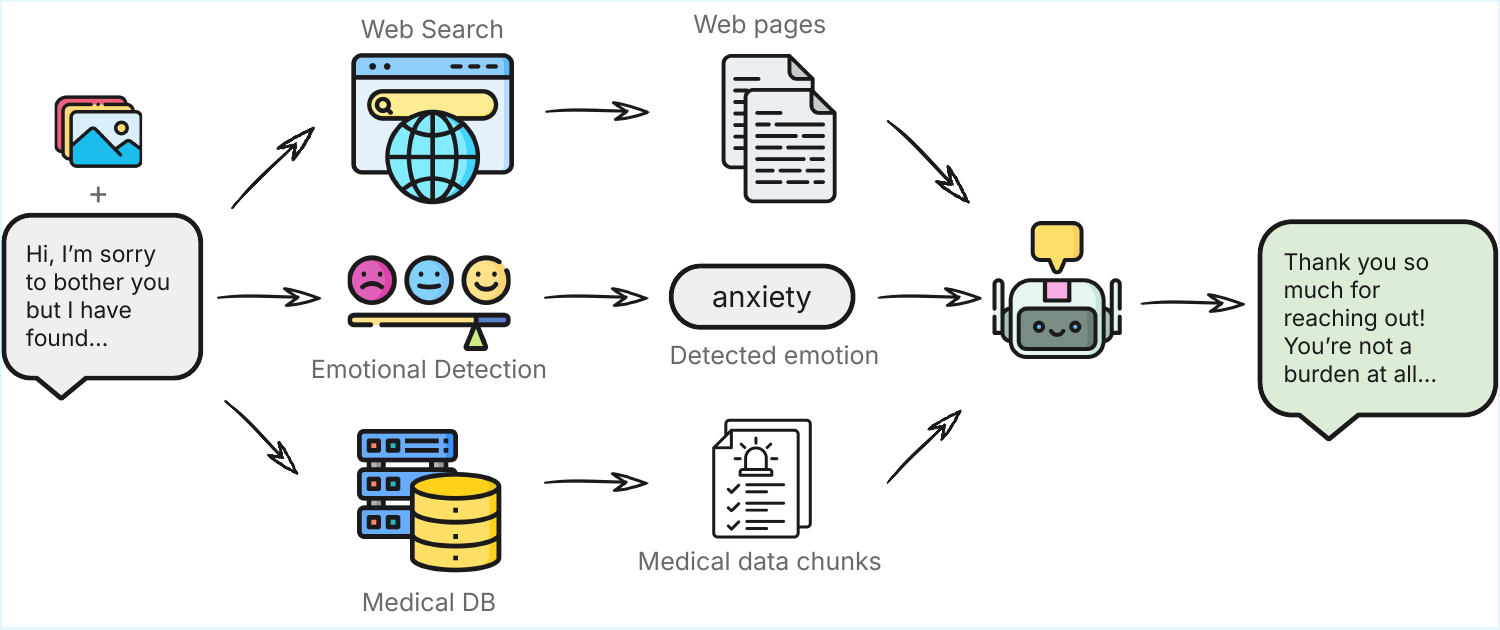}
\caption{Overview of the EmoMed architecture. A multimodal VLM detects user emotion, retrieves information from both a web search and a medical database, and generates empathetic, evidence-grounded responses. Emotion conditioning guides tone and phrasing, while dual retrieval ensures factual accuracy and recency in safety-constrained medical dialogue.}
\label{fig:overview}
\end{figure*}

Moving from bare LLMs to tool-using agents (function calls, web search, access to medical databases, and retrieval-augmented generation (RAG)) improves recency, grounding, and traceability of clinical information \citep{miao2024integratingRAGNephrology,fernandez2025check}. At the same time, safety becomes central: dynamic red-teaming reveals vulnerabilities in robustness, privacy, and cognitive-bias priming \citep{pan2025redteaming}, and healthcare-specific safety benchmarks (e.g., CARES) formalize principle-grounded risk evaluation \citep{chen2025cares}. Real-world, clinician-in-the-loop validation (RWE-LLM) further reduces severe harm and improves correctness \citep{bhimani2025rwellm}. Operationally, programmable guardrails are increasingly adopted for medical scenarios \citep{rebedea2023nemo,hakim2025guardrails}. Finally, data poisoning remains a core threat \citep{alber2025poisoning}, and stronger reasoning without proper constraints can amplify harm on unsafe inputs \citep{zhou2025hiddenRisksR1}.

Many clinical questions are image-grounded (e.g., dermatology, radiology, pathology), making multimodality a necessity. Vision–language models (VLMs / MLLMs) have advanced medical visual question answering (Med-VQA) and report generation \citep{moon2021multimodal,hartsock2024vision,yi2025radiology}. The ecosystem of datasets and benchmarks continues to expand (PMC-VQA, SLAKE, VQA-RAD, PathVQA) \citep{zhang2024pmcvqa,liu2021slake,lau2018vqarad,he2020pathvqa}, alongside methods to leverage external knowledge and reasoning-oriented prompting \citep{ma2025seek,lan2023improving}. Remaining challenges include limited clinically annotated data, reliance on external knowledge, and robust, safety-aware evaluation \citep{hartsock2024vision,yi2025radiology}.

LLMs/MLLMs increasingly automate support tasks: medical dialogue / document summarization, transcription and note generation, reference Q\&A, and educational content \citep{liu2024meddialogue,thomas2025ondevice,yang2024llmsynergy}. Biomedical benchmarks (e.g., BLURB) assess understanding and information extraction \citep{feng2024blurb}, while clinical QA benchmarks (MedQA, MultiMedQA) probe knowledge and reasoning \citep{singhal2022nature,singhal2023medpalm}. Despite progress, safe adoption requires transparent sourcing, explicit uncertainty communication, and robust error analysis \citep{alwakeel2025rigor}.

Affective personalization is key to acceptance and communication quality: emotionally adaptive and empathetic responses can improve engagement, perceived quality, and clarity -- sometimes preferred to physician responses in controlled settings \citep{ghandeharioun2019towards,ayers2023comparing,luo2024assessing,hua2025charting}. In healthcare, anxiety, confusion, and urgency shape risk perception, adherence, and trust; however, style control must never degrade factuality or safety. We therefore build on recent findings about emotion-adaptive interaction improving behavioral and perceptual outcomes in human–AI communication \citep{mozikov2024eai, mozikov2025hl}, and align these insights with clinically oriented safety and accuracy metrics \citep{heersmink2024phenomenology,concannon2023empathy,yuceercetiner2025orthognathic,lang2024scoliosis}.

We introduce a medical consultation agent that: (i) processes text and medical images; (ii) detects affect indicators (anxiety, confusion, urgency) and adapts tone, structure, and level of detail; and (iii) strengthens factual reliability via dual grounding: web search for up-to-date guidance and an API-connected, continuously updated medical knowledge base. We compare seven frontier LLMs (GPT-4o/5, Qwen3, Llama 4, Gemini 2.5, Grok 4, Claude 3) using comprehensive metrics: LLM-as-judge (empathy, clarity, helpfulness, safety, consistency, relevance), MedQA-style accuracy, semantic similarity (BERTScore), and multimodal medical benchmarks. Emotion-adaptive responses have been shown to consistently improve perceived empathy and communication clarity, without compromising clinical accuracy, compared to emotion-agnostic approaches. A controlled study conducted on real individuals has confirmed these findings.

\section{Related Work}

Conversational agents and LLMs are widely applied in healthcare for education, decision support, and summarization \citep{car2020conversational,barreda2025transforming,liu2025reviewApplyingLLMHealthcare}. Frontier models achieve high medical reasoning scores \citep{singhal2022nature,nori2023gpt4capabilities,singhal2023medpalm}, yet evaluations highlight limits in realism and transparency \citep{alwakeel2025rigor}. To curb hallucination and drift, retrieval-augmented and tool-using designs are common \citep{miao2024integratingRAGNephrology,fernandez2025check}, while safety research emphasizes red-teaming and domain-specific benchmarks \citep{pan2025redteaming,chen2025cares} and countermeasures against poisoning or overconfidence \citep{alber2025poisoning,zhou2025hiddenRisksR1}. Guardrails and abstention remain key mitigations \citep{rebedea2023nemo,hakim2025guardrails}.

Affective personalization improves empathy and engagement \citep{ghandeharioun2019towards,ayers2023comparing,hua2025charting}, using emotion recognition, multimodal cues, and style conditioning. Yet persona steering risks bias, motivating policy-level tone control in clinical contexts \citep{deshpande2023toxicity}. Because trust and comprehension depend on both tone and clarity, our work applies interpretable style parameters that adapt warmth, hedging, and structure to user affect while preserving factual accuracy and providing uncertainty and escalation cues.

Multimodal models advance medical VQA \citep{zhang2024pmcvqa,liu2021slake,lau2018vqarad,he2020pathvqa} but face grounding and safety challenges \citep{hartsock2024vision,yi2025radiology}. We employ dual retrieval (web + KB) to mitigate hallucination \citep{fernandez2025check,ozbay2025endodontics,jedrzejczak2025audiology}. Evaluation combines LLM-as-judge \citep{croxford2025llmasjudge,yu2025agentsAsJudges}, factuality checks, and semantic metrics \citep{zhang2019bertscore}, validated through user studies on empathy, clarity, and trust \citep{gu2024survey}.

\begin{figure}[!t]
\centering
\includegraphics[width=0.98\columnwidth]{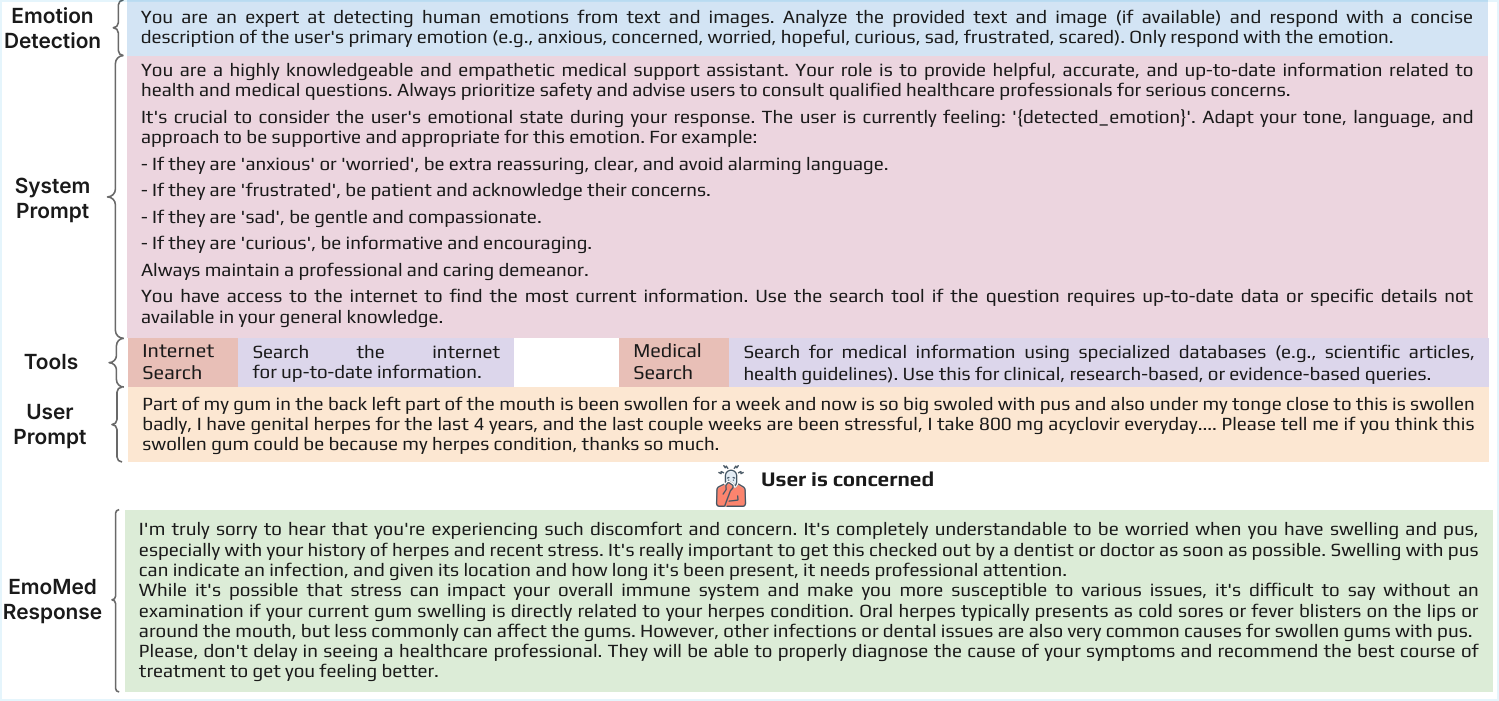}
\captionof{figure}{End-to-end EmoMed pipeline illustrating emotion detection, system prompting, dual retrieval tools (web and medical databases), and the generation of empathetic, clinically grounded responses. The example shows how detected emotion and retrieved evidence jointly condition tone, content, and guidance in a safety-aligned medical consultation.}
\label{fig:example}

\vspace{0.8em}

\begin{minipage}[t]{0.49\columnwidth}
\centering
\includegraphics[width=\linewidth]{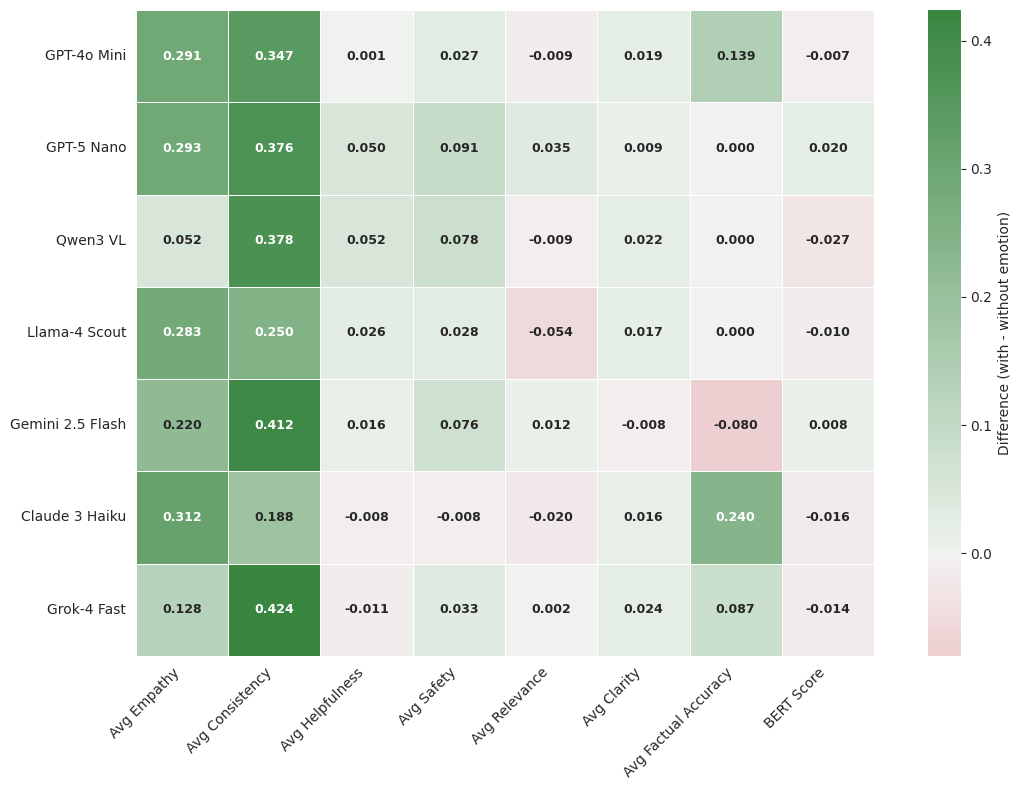}
\captionof{figure}{Text QA: heatmap of LLM-as-judge metrics across models (higher is better).}
\label{fig:heatmap-text}
\end{minipage}\hfill
\begin{minipage}[t]{0.49\columnwidth}
\centering
\includegraphics[width=\linewidth]{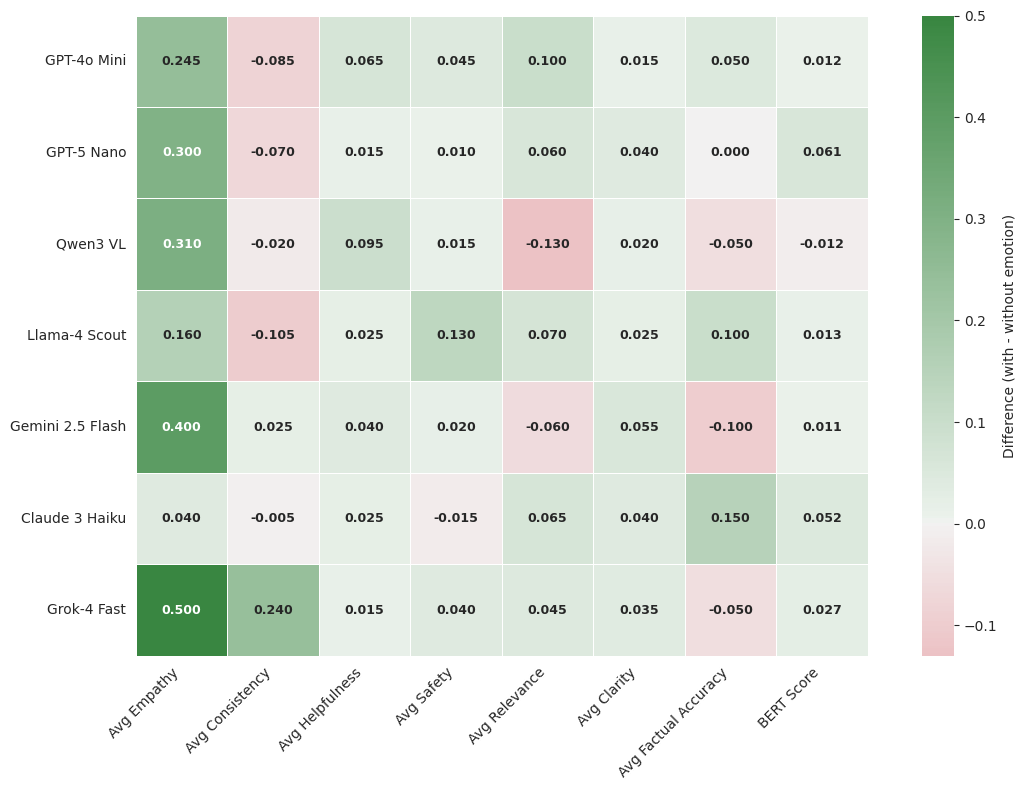}
\captionof{figure}{Med-VQA: heatmap of LLM-as-judge metrics across models (higher is better).}
\label{fig:heatmap-vqa}
\end{minipage}
\end{figure}

\section{Architecture}

Our system is a multimodal agent built around a VLM that processes user queries consisting of text and optional medical images. See Figure~\ref{fig:overview} for a high-level overview of the system. It detects the user’s emotional state, retrieves up-to-date clinical information from external sources, and generates empathetic, fact-grounded responses while adhering to strict safety constraints.

The agent operates in a tool-augmented loop: the VLM decides whether to answer directly or invoke one of two retrieval tools -- (1) a general web search Tavily for time-sensitive information (e.g., public health updates), or (2) a domain-specific medical knowledge base MediSearch~\cite{trivedi2023medisearch} for evidence-based clinical content. Tool calls are limited to five iterations per query to ensure efficiency and prevent runaway reasoning. A complete tool-call trace for a typical query is shown in Figure~\ref{fig:example}.

Emotion detection is handled by the GPT-4o model using a specialized instruction that directs it to infer the user’s primary emotional state from the input text and any accompanying image. The model is constrained to output only a single emotion term (such as \textit{anxious}, \textit{concerned}, \textit{hopeful}, or \textit{frustrated}), without additional explanation. If the output cannot be parsed as a valid emotion, the system continues with \textit{neutral} emotional context.

The response generation prompt instructs the VLM to: avoid diagnostic claims, acknowledge uncertainty, recommend professional care for serious concerns, and, when an emotion is detected, adjust tone to be supportive and appropriate. Responses follow a consistent structure: acknowledgment of the user’s concern, a brief restatement, relevant clinical context (without alarmism), actionable guidance, and, when helpful, a clarifying follow-up question.

We evaluate two versions of the agent:  \textbf{emotion-aware} --- uses the detected emotion to shape response tone and phrasing; \textbf{baseline} --- uses an identical prompt but with all emotion-related instructions removed (no emotion is passed or referenced).
For evaluation, we use two datasets. The text-only benchmark is drawn from \texttt{lavita/medical-qa-datasets}, which combines the Medical Meadow and ChatDoctor–HealthcareMagic collections of physician-style medical questions. For multimodal evaluation, we use \texttt{flaviagiammarino/path-vqa}, a dataset of visual question answering tasks based on pathology images from standard textbooks and the PEIR digital library~\cite{he2020pathvqa}.

Our system is designed to accommodate user-uploaded images and photographs, though the primary emotional context is conveyed through users' textual messages. The multimodal capabilities serve to support the consultation scenario by enabling the system to process visual information when provided by users, while maintaining focus on emotional understanding from text-based communication.

Automatic evaluation uses GPT-4o as a judge to assess factual accuracy (binary), empathy, emotional consistency, helpfulness, safety, relevance, and clarity on a 1–10 scale. We also compute semantic similarity between model and reference answers using Sentence-BERT. In addition, we conducted a user study with 48 participants who compared emotion-aware and baseline responses in a blinded pairwise setting and selected the version they found more appropriate and supportive. Results are reported in Section Results.

The implementation uses a single VLM endpoint (via OpenRouter) with temperature fixed at 0.1. All components (emotion detection, tool selection, and response generation) are handled within the same model call sequence, ensuring a lightweight and reproducible pipeline.

\section{Results}
\label{sec:results}

\paragraph{Overall Effects. }
Across seven vision-capable LLMs, emotion-aware prompting reliably improved affective dimensions while preserving core correctness. On text QA, empathy and clarity increased across all models, with safety and helpfulness broadly stable and small, mixed changes in BERTScore and relevance. On Med-VQA, the pattern held, empathy and clarity improved, safety trended up, while factual accuracy remained mixed given higher task difficulty. Figures~\ref{fig:heatmap-text}--\ref{fig:radar-vqa} visualize metric distributions and the margin over the neutral baseline.

\paragraph{Text QA (aggregate). } 
Table~\ref{tab:text-results} reports eight metrics with and without emotion-aware prompting. Typical gains include +0.20-0.35 in empathy and +0.01-0.03 in clarity; safety shifts are small and positive on average. Accuracy is unchanged or higher for most models (e.g., GPT-4o Mini +0.139, Grok-4 Fast +0.087), with one notable decrease on Gemini 2.5 Flash (-0.080), despite strong affect gains.

\paragraph{Key Findings.} Affective improvements were achieved without detectable content drift: empathy and clarity increased across all models, while semantic similarity (BERTScore) and relevance remained close to baseline. Accuracy was predominantly stable or higher. Four models matched or exceeded baseline accuracy, whereas Gemini 2.5 exhibited a modest accuracy decrement in exchange for substantial affective gains.

\paragraph{Med-VQA (aggregate).}
Table~\ref{tab:vqa-results} mirrors the text QA pattern on image-grounded tasks. Empathy increases are robust across models, clarity improves modestly, and safety trends positive. Accuracy differences are small and mixed (e.g., Llama-4 Scout +0.101; Gemini 2.5 -0.098). Emotional consistency can dip in some VQA settings, likely due to noisier affect cues in image prompts.

\paragraph{Text QA: visual summary.}
Empathy and clarity consistently increase with emotion-aware prompting, while correctness-linked metrics remain stable (Figures~\ref{fig:heatmap-text}--\ref{fig:radar-text}).

\paragraph{Med-VQA: visual summary.}
Trends mirror text QA - affective gains with stable content metrics—though accuracy is lower overall due to task difficulty (Figures~\ref{fig:heatmap-vqa}--\ref{fig:radar-vqa}).

\paragraph{User Study.}
In blinded pairwise comparisons, 44 participants aged 19-37 (19 women, 25 men) preferred emotion-aware responses for empathy and communication clarity without perceiving losses in factual accuracy or safety. Of these participants, 34 (77\%) rated the emotional response as better than the alternative. Free-text feedback highlighted clearer structure for confusion, reassurance for anxiety, and calibrated hedging for urgent cases.

\paragraph{Error Analysis and Failure Modes.}
Instances of over-hedging and verbosity were rare, where excessive caveats sometimes reduced directness; we mitigate this with a fixed content plan and caps on detail. Emotional misclassification occurred occasionally on brief or ambiguous inputs, and defaulting to a neutral‑warm tone limits the downside. Tool-use and grounding misses were infrequent, typically stale retrieval or extraction misses - and are addressed through retries, transparent sourcing, and escalation language. In VQA, weaker affect cues in images can lead to drift and reduce emotional consistency, so we bias toward concise, neutral‑warm responses. Overall, these issues were uncommon and did not change aggregate trends; affective gains persisted even after filtering out low-confidence affect detections.

\section{Conclusion and Future Work}
EmoMed combines emotion-aware style control with dual retrieval to deliver patient-centered responses without sacrificing rigor. Across seven models and both text and image QA, emotion-aware prompting consistently improved empathy and clarity while keeping factual accuracy, safety, and relevance on par with (or better than) a neutral baseline; a blinded user study corroborated these benefits.

% Limitations include single-label affect modeling, LLM-as-judge biases, and curated datasets.
Future work will target richer affect dynamics, stronger safety and provenance, specialty-grounded retrieval, and larger human studies in clinical workflows.

\section{Limitations}
The paper acknowledges several limitations of the study: first, the emotion model relies on a simplified, single-label approach, which may not fully capture the complexity and subtlety of human emotions; second, automatic evaluation using an LLM as a judge may introduce systematic bias due to the subjective nature of language models; third, the experiments were conducted using curated datasets, which may limit the generalizability of the results to broader, more realistic scenarios. The authors also recognize the need for further research involving larger groups of users and integration into clinical practices.

% \section{Acknowledgements}
% The work of I. Makarov was supported by the Ministry of Economic Development of the Russian Federation (agreement No. 139-10-2025-034 dd. 19.06.2025, IGK 000000C313925P4D0002).

\appendix
\section{Appendix A. Average metrics for method across all models}

\begin{figure}[H]
\centering
\begin{minipage}[t]{0.45\columnwidth}
\centering
\includegraphics[width=0.95\linewidth]{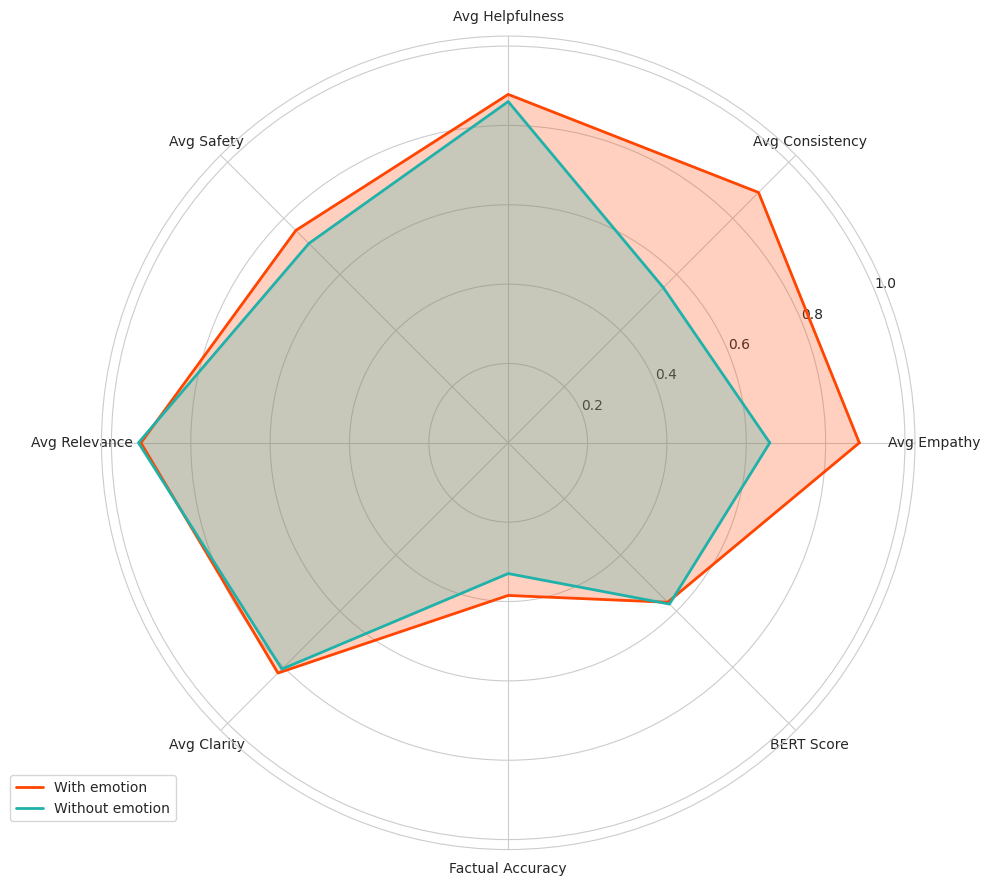}
\captionof{figure}{Text QA: radar chart for emotion-aware vs.\ baseline (empathy, clarity, helpfulness, safety).}
\label{fig:radar-text}
\end{minipage}\hfill
\begin{minipage}[t]{0.45\columnwidth}
\centering
\includegraphics[width=0.95\linewidth]{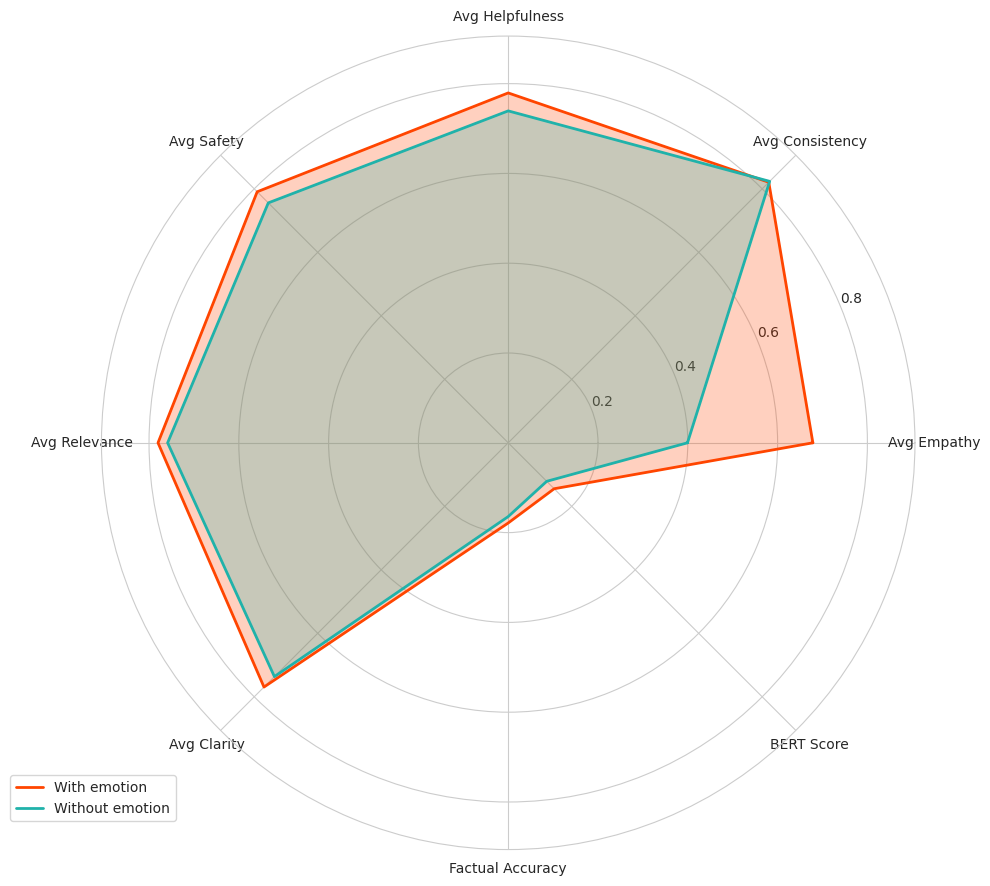}
\captionof{figure}{Med-VQA: radar chart for emotion-aware vs.\ baseline.}
\label{fig:radar-vqa}
\end{minipage}
\end{figure}

Figures~\ref{fig:radar-text} and~\ref{fig:radar-vqa} present radar charts comparing the emotion-aware agent and the baseline across key dimensions for text-based and multimodal medical QA, respectively. In both settings, the emotion-aware variant matches or slightly exceeds the baseline on factual and safety metrics while achieving substantially higher empathy scores.

\section{Appendix B. Detailed metrics for individual models}

Appendix B provides detailed performance metrics for individual models evaluated with and without emotional prompting on text-only and VQA benchmarks. Table~\ref{tab:text-results} presents comprehensive results for the text-only benchmark, showing that emotional prompting consistently improves metrics across all six models. Table~\ref{tab:vqa-results} demonstrates similar patterns for the VQA benchmark, with emotional prompting particularly enhancing empathy and helpfulness scores, though consistency metrics tend to be higher without emotional prompting in this modality.

\begin{table*}[!t]
\centering
\small
\begin{minipage}[t]{0.49\textwidth}
\centering
\captionof{table}{Comparison of Model Performance with and without Emotional Prompting across Two Benchmarks (text only). FA --- Factual Accuracy, AE - Avg Empathy, ACon --- Avg Consistency, AH --- Avg Helpfulness, AS --- Avg Safety, BS --- BERT Score, AR --- Avg Relevance, AClar --- Avg Clarity}
\label{tab:text-results}
\adjustbox{width=\linewidth,center}{%
\begin{tabular}{llcc}
\toprule
\textbf{Model} & \textbf{Metric} & \textbf{\makecell{With Emotion}} & \textbf{\makecell{Without Emotion}} \\
\midrule

% === TEXT BENCH ===
\multirow{2}{*}{GPT-4o Mini} & FA & \best{0.457} & 0.318 \\
& AE & \best{0.919} & 0.628 \\
& ACon & \best{0.923} & 0.576 \\
& AH & \best{0.858} & 0.857 \\
& AS & \best{0.718} & 0.691 \\
& BS & 0.579 & \best{0.586} \\
& AR & 0.936 & \best{0.945} \\
& AClar & \best{0.829} & 0.810 \\
\cmidrule(lr){1-4}

\multirow{2}{*}{GPT-5 Nano} & FA & 0.391 & 0.391 \\
& AE & \best{0.900} & 0.607 \\
& ACon & \best{0.913} & 0.537 \\
& AH & \best{0.907} & 0.857 \\
& AS & \best{0.498} & 0.407 \\
& BS & \best{0.541} & 0.521 \\
& AR & \best{0.930} & 0.896 \\
& AClar & \best{0.813} & 0.804 \\
\cmidrule(lr){1-4}

\multirow{2}{*}{Qwen3 VL} & FA & 0.239 & 0.239 \\
& AE & \best{0.857} & 0.804 \\
& ACon & \best{0.857} & 0.478 \\
& AH & \best{0.885} & 0.833 \\
& AS & \best{0.824} & 0.746 \\
& BS & 0.500 & \best{0.527} \\
& AR & 0.896 & \best{0.904} \\
& AClar & \best{0.833} & 0.811 \\
\cmidrule(lr){1-4}

\multirow{2}{*}{Llama-4 Scout} & FA & 0.196 & 0.196 \\
& AE & \best{0.796} & 0.513 \\
& ACon & \best{0.846} & 0.596 \\
& AH & \best{0.826} & 0.800 \\
& AS & \best{0.791} & 0.763 \\
& BS & 0.560 & \best{0.570} \\
& AR & 0.822 & \best{0.876} \\
& AClar & \best{0.820} & 0.802 \\
\cmidrule(lr){1-4}

\multirow{2}{*}{Gemini 2.5 Flash} & FA & 0.480 & \best{0.560} \\
& AE & \best{0.940} & 0.720 \\
& ACon & \best{0.916} & 0.504 \\
& AH & \best{0.904} & 0.888 \\
& AS & \best{0.912} & 0.836 \\
& BS & \best{0.611} & 0.603 \\
& AR & \best{0.964} & 0.952 \\
& AClar & 0.808 & \best{0.816} \\
\cmidrule(lr){1-4}

\multirow{2}{*}{Claude 3 Haiku} & FA & \best{0.560} & 0.320 \\
& AE & \best{0.820} & 0.508 \\
& ACon & \best{0.840} & 0.652 \\
& AH & 0.868 & \best{0.876} \\
& AS & 0.748 & \best{0.756} \\
& BS & 0.603 & \best{0.619} \\
& AR & 0.952 & \best{0.972} \\
& AClar & \best{0.816} & 0.800 \\
\cmidrule(lr){1-4}

\multirow{2}{*}{Grok-4 Fast} & FA & \best{0.370} & 0.283 \\
& AE & \best{0.963} & 0.835 \\
& ACon & \best{0.952} & 0.528 \\
& AH & 0.900 & \best{0.911} \\
& AS & \best{0.809} & 0.776 \\
& BS & 0.584 & \best{0.598} \\
& AR & \best{0.980} & 0.978 \\
& AClar & \best{0.826} & 0.802 \\

\bottomrule
\end{tabular}
}
\end{minipage}\hfill
\begin{minipage}[t]{0.49\textwidth}
\centering
\captionof{table}{Comparison of Model Performance with and without Emotional Prompting across Two Benchmarks (VQA). FA --- Factual Accuracy, AE - Avg Empathy, ACon --- Avg Consistency, AH --- Avg Helpfulness, AS --- Avg Safety, BS --- BERT Score, AR --- Avg Relevance, AClar --- Avg Clarity}
\label{tab:vqa-results}
\adjustbox{width=\linewidth,center}{%
\begin{tabular}{llcc}
\toprule
\textbf{Model} & \textbf{Metric} & \textbf{\makecell{With Emotion}} & \textbf{\makecell{Without Emotion}} \\
\midrule

\multirow{2}{*}{GPT-4o Mini}
 & FA & \best{0.103} & 0.052 \\
& AE & \best{0.664} & 0.417 \\
& ACon & 0.732 & \best{0.818} \\
& AH & \best{0.743} & 0.678 \\
& AS & \best{0.792} & 0.748 \\
& BS & \best{0.109} & 0.099 \\
& AR & \best{0.694} & 0.593 \\
& AClar & \best{0.768} & 0.752 \\
\cmidrule(lr){1-4}

\multirow{2}{*}{GPT-5 Nano}
 & FA & 0.201 & \best{0.203} \\
 & AE & \best{0.682} & 0.383 \\
& ACon & 0.783 & \best{0.851} \\
& AH & \best{0.848} & 0.832 \\
 & AS & \best{0.777} & 0.768 \\
& BS & \best{0.152} & 0.091 \\
& AR & \best{0.873} & 0.813 \\
& AClar & \best{0.728} & 0.687 \\
\cmidrule(lr){1-4}

\multirow{2}{*}{Qwen3 VL}
& FA & 0.252 & \best{0.303} \\
& AE & \best{0.693} & 0.382 \\
& ACon & 0.823 & \best{0.842} \\
& AH & \best{0.913} & 0.818 \\
& AS & \best{0.768} & 0.753 \\
& BS & 0.174 & \best{0.186} \\
& AR & 0.863 & \best{0.993} \\
& AClar & \best{0.813} & 0.792 \\
\cmidrule(lr){1-4}

\multirow{2}{*}{Llama-4 Scout}
& FA & \best{0.153} & 0.052 \\
& AE & \best{0.588} & 0.428 \\
& ACon & 0.709 & \best{0.813} \\
& AH & \best{0.668} & 0.643 \\
& AS & \best{0.828} & 0.698 \\
& BS & \best{0.115} & 0.102 \\
& AR & \best{0.643} & 0.573 \\
& AClar & \best{0.778} & 0.754 \\
\cmidrule(lr){1-4}

\multirow{2}{*}{Gemini 2.5 Flash}
& FA & 0.203 & \best{0.301} \\
& AE & \best{0.773} & 0.374 \\
& ACon & \best{0.913} & 0.888 \\
& AH & \best{0.813} & 0.773 \\
 & AS & \best{0.728} & 0.707 \\
& BS & \best{0.199} & 0.190 \\
& AR & 0.788 & \best{0.849} \\
 & AClar & \best{0.787} & 0.735 \\
\cmidrule(lr){1-4}

\multirow{2}{*}{Claude 3 Haiku}
& FA & \best{0.203} & 0.052 \\
 & AE & \best{0.503} & 0.462 \\
& ACon & 0.888 & \best{0.893} \\
&AH & \best{0.653} & 0.628 \\
 & AS & 0.773 & \best{0.788} \\
& BS & \best{0.159} & 0.107 \\
& AR & \best{0.723} & 0.658 \\
& AClar & \best{0.743} & 0.702 \\
\cmidrule(lr){1-4}

\multirow{2}{*}{Grok-4 Fast}
& FA & 0.153 & \best{0.201} \\
& AE & \best{0.868} & 0.367 \\
& ACon & \best{0.918} & 0.678 \\
& AH & \best{0.838} & 0.823 \\
& AS & \best{0.889} & 0.847 \\
& BS & \best{0.121} & 0.094 \\
& AR & \best{0.898} & 0.853 \\
& AClar & \best{0.789} & 0.756 \\

\bottomrule
\end{tabular}
}
\end{minipage}
\end{table*}

\appendix

\bibliographystyle{splncs04}
\bibliography{aaai2026}





% LNCS Homepage, \url{http://www.springer.com/lncs}, last accessed 2023/10/25
% \end{thebibliography}
\end{document}